\documentclass{article}

\PassOptionsToPackage{numbers, compress}{natbib}

\usepackage[preprint]{nips_2018}

\usepackage[utf8]{inputenc} 
\usepackage[T1]{fontenc}    
\usepackage{hyperref}       
\usepackage{url}            
\usepackage{booktabs}       
\usepackage{amsfonts}       
\usepackage{nicefrac}       
\usepackage{microtype}      
\usepackage{graphicx}
\usepackage{float}
\usepackage{amsmath}
\usepackage{subfig}

\title{Neural Arithmetic Expression Calculator}

\author{Kaiyu Chen, Yihan Dong, Xipeng Qiu\thanks{Corresponding Author, xpqiu@fudan.edu.cn}, Zitian Chen\\
 Shanghai Key Laboratory of Intelligent Information Processing, Fudan University\\
School of Computer Science, Fudan University\\
\{15307130233, 15302010054, xpqiu, ztchen13\}@fudan.edu.cn
}

\begin{document}

\maketitle

\begin{abstract}
This paper presents a pure neural solver for arithmetic expression calculation (AEC) problem. Previous work utilizes the powerful capabilities of deep neural networks and attempts to build an end-to-end model to solve this problem. However, most of these methods can only deal with the additive operations. It is still a challenging problem to solve the complex expression calculation problem, which includes the adding, subtracting, multiplying, dividing and bracketing operations. In this work, we regard the arithmetic expression calculation as a hierarchical reinforcement learning problem. An arithmetic operation is decomposed into a series of sub-tasks, and each sub-task is dealt with by a skill module. The skill module could be a basic module performing elementary operations, or interactive module performing complex operations by invoking other skill models. With curriculum learning, our model can deal with a complex arithmetic expression calculation with the deep hierarchical structure of skill models. Experiments show that our model significantly outperforms the previous models for arithmetic expression calculation.
\end{abstract}

\section{Introduction}

\begin{figure}[H]
\centering
\includegraphics[width=0.56\linewidth]{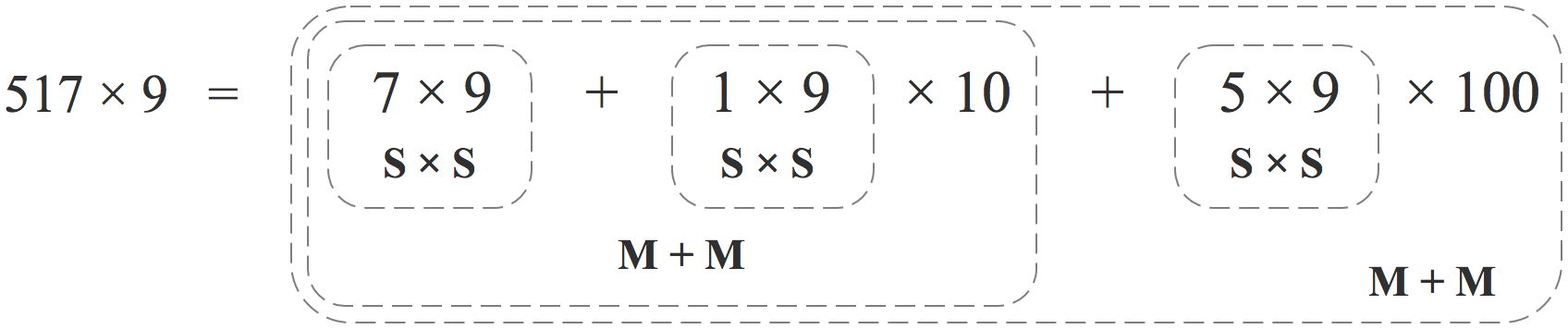}
\caption{An example of reusability and hierarchy in multiplication. $M+M$ denotes multi-digit addition, while $S\times S$  means single-digit multiplication.}\label{fig:example}
\end{figure}

Developing pure neural models to automatically
solve arithmetic expression calculation (AEC) is an interest and challenging task.
Recent research includes Neural GPUs~\citep{kaiser2015neural,freivalds2017improving}, Grid LSTM \citep{Kalchbrenner2015GridLS}, Neural Turing Machines \citep{graves2014neural}, and Neural Random-Access Machines~\citep{Kurach2016NeuralRM}. Most of these models just can deal with the addition calculation. Although Neural GPU has an ability to learn multi-digit binary multiplication, it does not work well in decimal multiplication \citep{kaiser2015neural}. The difficulty of multi-digit decimal multiplication lies in the fact that multiplication involves a complicated structure of arithmetic operations, which is hard for neural networks to learn.  Considering how electronic circuit or human beings do multiplication, multi-digit multiplication can be decomposed into several subgoals, as shown in Figure \ref{fig:example}. High-level arithmetic tasks like multiplication iteratively use low-level operations like the addition to complete high-level tasks.
The incapability of current models in solving arithmetic expression is because they fail to use two key properties of arithmetic operation: \textbf{reusability} and \textbf{hierarchy}. The arithmetic operation can be decomposed into a series of sub-operations, which form a hierarchical structure. Most of the sub-operations are reusable. When dealing with a complex arithmetic operation, we do not need to train a model from scratch. For the example in Figure  \ref{fig:example}, the multi-digit multiplication involves several reusable sub-operations, such as $S\times S$ and $M + M$.

To leverage reusability and hierarchy in the arithmetic operation, we formulate this task as a Hierarchical Reinforcement Learning (HRL) problem~\citep{Sutton1999BetweenMA,Dietterich2000HierarchicalRL}, in which the task policy can be decomposed into several sub-task policies. Each sub-task policy is implemented by a \textit{skill module}, which can be used recursively. The skill module can be divided into two groups: basic skill module performing elementary single-digit operations, and interactive skill model performing complex operations by selectively invoking other skill modules.
There are two differences to the standard HRL.  (1) One is that each invoked skill module can be executed with only its input, regardless of external environment state. Therefore, we propose \textbf{Interactive Skill Modules (ISM)} that can selectively interact with other skill modules by sending a partial expression and receiving answers returned. (2) Another is that the task hierarchy is multi-level, which is difficult to be learned from scratch. Therefore, we propose \textbf{Curriculum Teacher and Continual-learning Student (CTCS)} framework to overcome this problem. The skill modules are trained in a particular order, from easy to difficult tasks. The finally skill module would be  a deep hierarchical structure.
The experiments show that our model has a strong capability to calculate arithmetic expressions.

%
%



The main contributions of the paper are:
\begin{itemize}
\item We propose a pure neural model to solve the (decimal) expression calculation problem, involving the $\{+, -, \times, \div, (, )\}$ operations. Both the input and output of our model is character sequence. To the best of our knowledge, this study is the first work to solve this challenging problem.
\item We regard arithmetic learning as a Multi-level Hierarchical Reinforcement Learning (MHRL) problem, and factorize a complex arithmetic operation into several simpler operations. The main component is the interactive skill module. A high-level interactive skill module can invoke the low-level skill modules by sending and receiving messages.
\item We introduce Curriculum Teacher and Continual-learning Student (CTCS), an automatic teacher-student framework that enables the model to be easier learned for the complex tasks.
\end{itemize}

\section{Related Work}
\paragraph{Arithmetic Learning}
In recent years, several models have attempted to learn arithmetic in deep learning. Grid LSTM \cite{Kalchbrenner2015GridLS} expands LSTM in multiple dimensions and can learn multi-digit addition. \citet{Zaremba2016LearningSA} use reinforcement learning to learn single-digit multiplication and multi-digit addition. Neural GPU \cite{kaiser2015neural} is noticeably promising in arithmetic learning and can learn binary multiplication. \citet{Price2016ExtensionsAL} and \citet{freivalds2017improving} improve Neural GPU to do multi-digit multiplication with curriculum learning. Nevertheless, there is no successful attempt to learn division or expression calculation.
\paragraph{Hierarchical Reinforcement Learning}

The first popular hierarchical reinforcement learning model may date back to the options framework \cite{Sutton1999BetweenMA}. The options framework considers the problem to have a two-level hierarchy. Recent work combines neural networks with this two-level hierarchy and has made promising results in challenging environments with sparse rewards, like Minecraft \cite{tessler2017deep} and ATARI games \cite{Baranes2013ActiveLO}. In contrast to the two-level hierarchy, the skill modules in our framework can selectively use other skill modules, which finally form a deep multi-level hierarchical structure.


\paragraph{Curriculum Learning}

Work by \citet{Bengio2009CurriculumL} brings general interests to curriculum learning. Recently, it has been widely used in many tasks, like learning to play first-person shooter games \cite{Wu2017TrainingAF}, and helping robots learn object manipulation \cite{Baranes2013ActiveLO}. It is noteworthy that the teacher-student curriculum learning framework proposed by \citet{matiisen2017teacher} can automatically sample tasks according to student's performance. However, it is limited to sampling data and can not help the student adapt to task switching with parameter adjustment.

\paragraph{Continual Lifelong Learning}

As proposed in \cite{tessler2017deep}, a continual lifelong learning model needs the ability to choose relevant prior knowledge for solving new tasks, which is named \textbf{selective transfer}. The main issue of continual learning models is that they are prone to \textbf{catastrophic forgetting} \cite{Mcclelland1995WhyTA,Parisi2018ContinualLL}, which means the model forgets previous knowledge when learning new tasks. To achieve continual lifelong learning, Progressive Neural Networks (PNN) \cite{Rusu2016ProgressiveNN} allocate a new module with access to prior knowledge to learn a new task. With this approach, prior knowledge can be used, and former modules are not influenced. Our model extends PNN with the ability to use helpful modules selectively.


\section{Model}

\paragraph{Task Definition}
We first formalize the task of arithmetic expression calculation (AEC) as follows. Given a character sequence, consisting of decimal digits $[0-9]$ and arithmetic operators of $\{+, -, \times, \div, (, )\}$, the goal is to output a sequence of digit characters representing the result, for example:
\begin{center}
\centering
\includegraphics[width=0.78\linewidth]{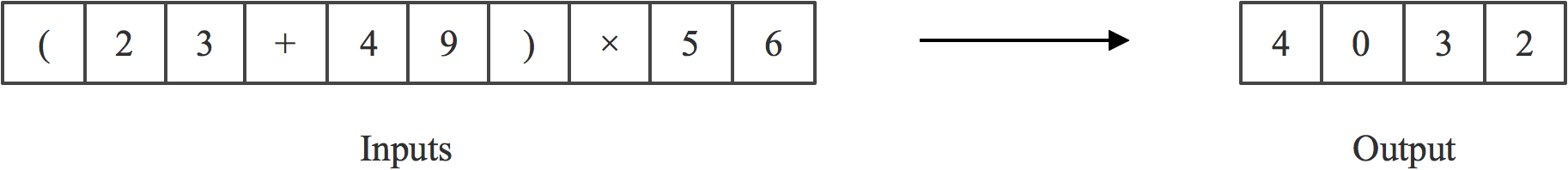}
\end{center}

\subsection{Multi-level Hierarchical Reinforcement Learning}

As analyzed before, the arithmetic calculation can be decomposed into several sub-tasks, including single-digit multiplication, multi-digit addition and more.
Assuming we already have several modules for the simple arithmetic calculations, the key challenge is how to organize them to solve a more complex arithmetic calculation.
In this paper, we propose a multi-level hierarchical reinforcement learning (MHRL) framework to perform this task.

\paragraph{Hierarchical Reinforcement Learning (HRL)}
In HRL, the policy $\pi$ of an agent can be decomposed into several sub-policies from the set $\Pi=\{\pi_1,\pi_2,\cdots,\pi_K\}$. At time $t$, the policy $\pi: \mathcal{S}\rightarrow \Pi$ is a mapping from state $s^t\in \mathcal{S}$ to a probability distribution over sub-policies. Assuming the $k$-th sub-policy is chosen, the action $a_t$ is determined by $\pi_k(a|s^t)$.

The arithmetic calculation is a multi-level hierarchical reinforcement learning, in which the sub-policy could be further decomposed into sub-sub-policies. Suppose that each (sub-)policy is implemented by a skill module. There are two different kinds of modules: \textit{basic skill modules} (BSM) and \textit{interactive skill modules} (ISM). All modules use character sequences as inputs and produce character sequences as outputs.

\begin{figure}[h]\centering
\subfloat[Basic Skill Module (BSM)]{\label{fig:basic}
    \includegraphics[width=0.325\textwidth]{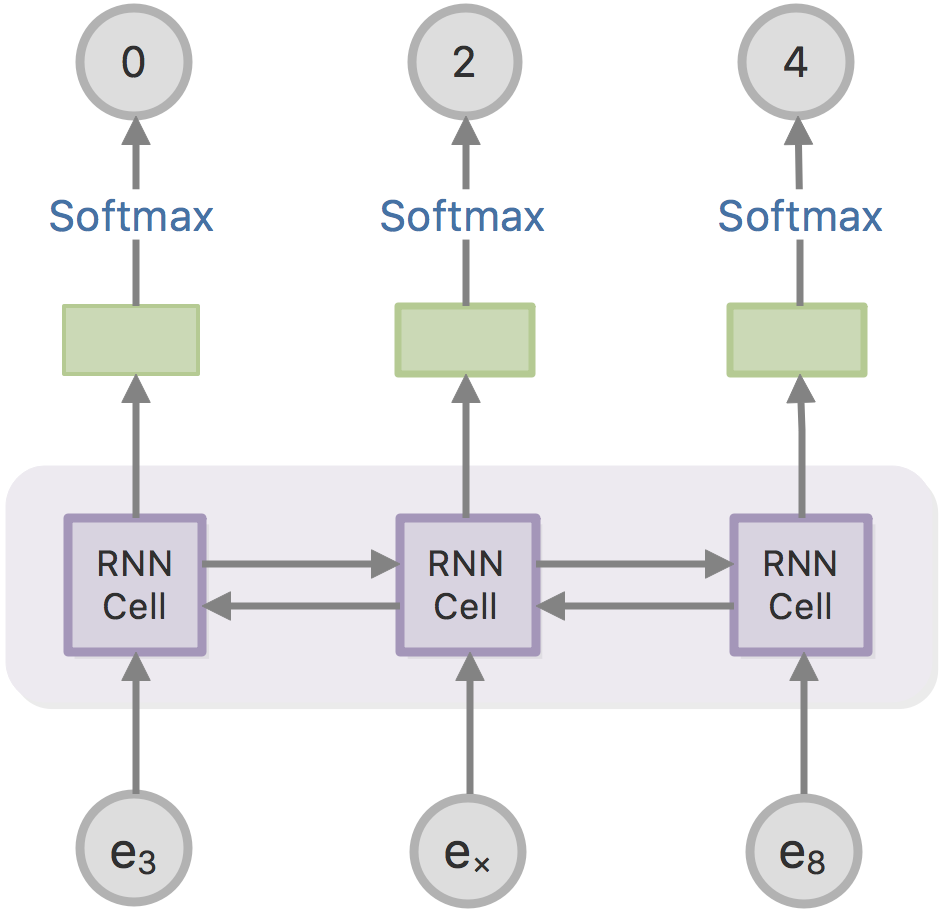}
}
\subfloat[Interactive Skill Module (ISM)]{\label{fig:sampleISM} \includegraphics[width=0.5\textwidth]{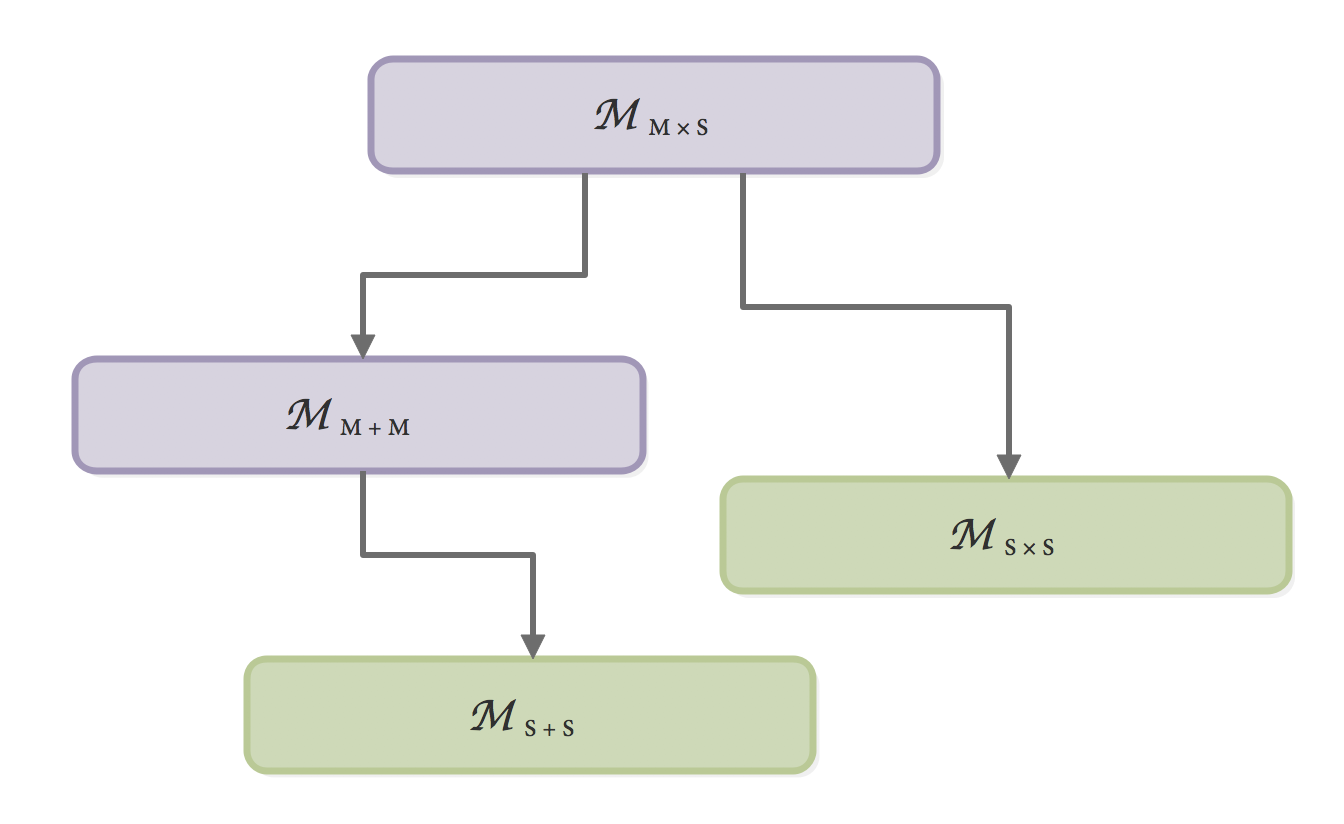}
}
\caption{Demonstration of skill modules. (a) the basic skill module (BSM) is a neural network with Bi-RNN, which takes the character sequence as input and outputs the character sequence of the result. (b) The interactive skill module (ISM) interacts with other skill models to do the calculation. This illustrated module $\mathcal{M}_{M \times S}$ can perform multi-digit $\times$ single-digit calculation, using two BSMs $\mathcal{M}_{S + S}$ (single-digit addition), $\mathcal{M}_{S \times S}$ (single-digit multiplication) and another ISM $\mathcal{M}_{M + M}$ for multi-digit addition.}\label{fig:sample}
\end{figure}

\paragraph{Basic Skill Modules (BSM)}

The basic skill modules perform fundamental arithmetic operations like single-digit's addition or multiplication. The structure of basic skill modules is illustrated in Figure \ref{fig:basic}. Given a sequence $c_1, c_2, \cdots, c_l$ containing decimal and arithmetic characters of length $l$. We firstly map the sequence with character embeddings to $e_{c_1}, e_{c_2}, \cdots, e_{c_l}$. Then the inputs are fed into a bi-directional RNN (Bi-RNN). Outputs are generated by choosing characters with the max probability after $\mathrm{softmax}$ functions. Basic skill modules are trained in a supervised approach.

Each BSM provides a deterministic policy $a_{1:l}=\pi(c_{1:l})$, where $a_{1:l}$ is the calculated result in form of a digit character sequence.

\paragraph{Interactive Skill Modules (ISM)}

The interactive skill modules perform the arithmetic operations by invoking other skill modules. An example of interactive skill modules is shown in Figure \ref{fig:sampleISM}.
The policy of ISM is to select other skill modules to complete the partial arithmetic calculation.
Different from the standard HRL, each skill module performs a local arithmetic operation, and need not observe the global environment state. Therefore, when a skill module $\mathcal{M}_i$ chooses another skill module $\mathcal{M}_j$ as sub-policy, module $\mathcal{M}_i$ just sends character sequences to module $\mathcal{M}_j$  and receives character sequences as answers.

\begin{figure}[t]
\centering
\includegraphics[width=0.8\linewidth,height=17em]{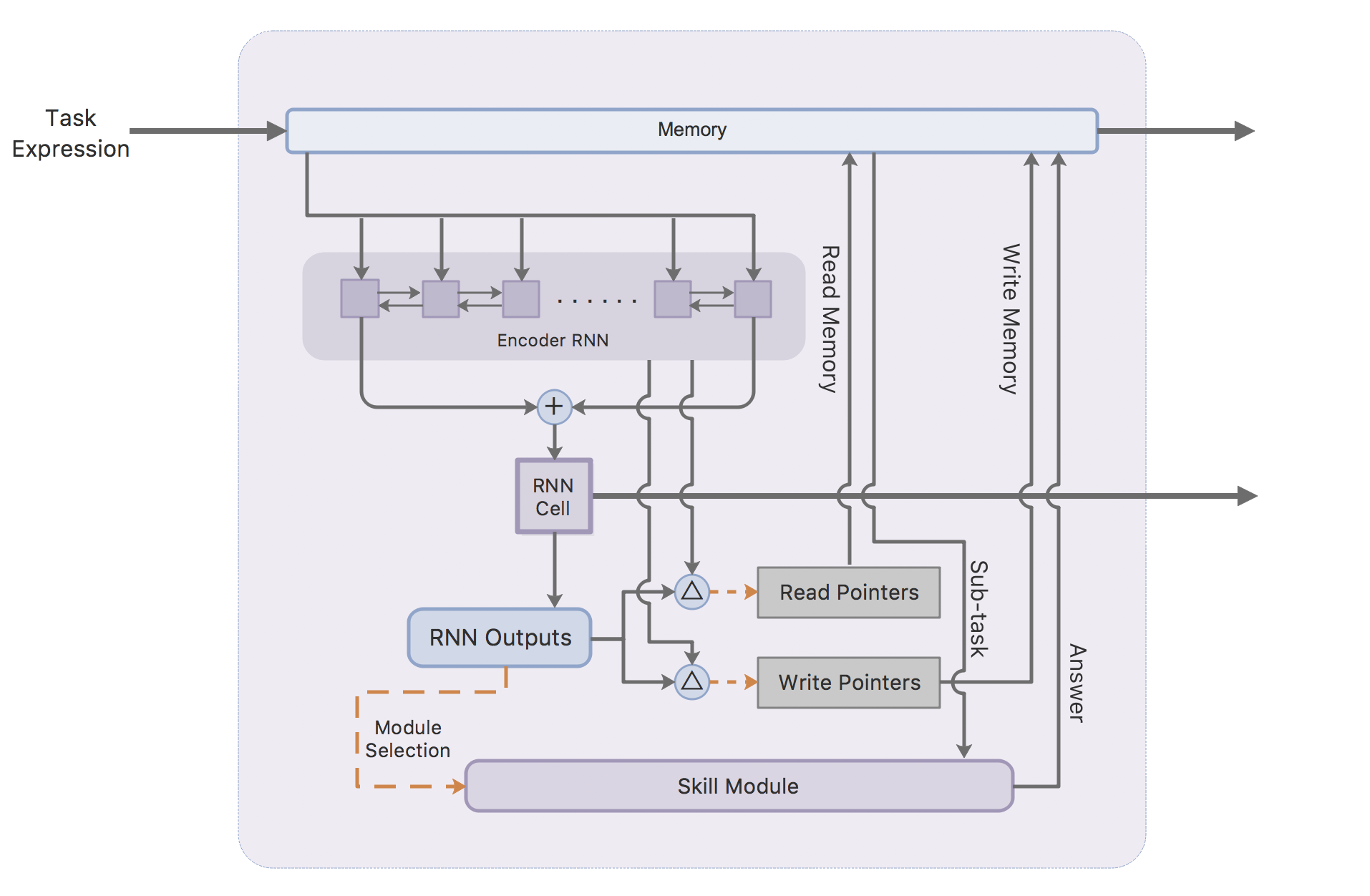}
\caption{Detailed structure of Interactive Skill Modules (ISM). In each time stamp, a bi-RNN encodes memory and output $o_1, o_2, \cdots, o_l$. The first and last outputs are concatenated as memory representations and are fed into the central RNN. The outputs of central RNN are used to select modules, generate read pointers and write pointers. The read pointers are used to read the sub-expression from memory. The selected module deal with the sub-expression and returns an answer. Then the answer is written to memory at positions indicated by write pointers. The dashed orange lines represent processes that are related to the policy of ISM.}\label{fig:memory}
\end{figure}


It is hard to train skill modules from scratch, so we use curriculum learning, which will be described in Section \ref{sec:curri}, to train skill modules in the order of increasing difficulty. Suppose that we already have $i-1$ well-trained skill modules $\Omega_\mathcal{M}=\{\mathcal{M}_1,\mathcal{M}_2,\cdots,\mathcal{M}_{i-1}\}$, the $i$-th ISM $\mathcal{M}_i$ is described as follows.

\subsubsection{Structure of Interactive Skill Module}

The detailed structure of ISM is shown in Figure \ref{fig:memory}.

First, each ISM is equipped with a memory $\xi$ to hold temporary information. Memory is composed of character slots with length $l$. When module $\mathcal{M}_i$ receives an expression $c_{1:l}$ , $\mathcal{M}_i$ first stores $c_{1:l}$ into memory.


The policy of ISM can be decomposed into three sub-policies: (1) selecting skill module, (2) reading memory, and (3) writing memory.

 At time $t$, the memory contains characters $c_1^t, c_2^t, \cdots, c_l^t$, we first use a Bi-RNN to encode the state of memory.
 \begin{align}
 (o_1^t, o_2^t, \cdots, o_l^t) &= \mathrm{Bi\textrm{-}RNN}(e_{c_1^t}, e_{c_2^t}, \cdots, e_{c_l^t}),
\end{align}
where $e_{c_k^t}$ is the embedding of character $c_k^t$ for $1\leq k \leq l$.

The state $s^t$ of the environment is modeled by a forward RNN,
\begin{align}
 h^t = \mathrm{RNNCell}[o_1^t\oplus o_l^t, h^{t-1}], \qquad
 s^t = \mathrm{FNN}(h^t),\label{eq:memory1}
\end{align}
where $\mathrm{FNN}(\cdot)$ is one-layer forward neural network.

Given the state $s^t$, the agent chooses three actions according the three following sub-policies,
\begin{align}
 \mathcal{M}^t &= \mathrm{ModuleSelect}([\mathcal{M}_1, \mathcal{M}_2, \cdots, \mathcal{M}_{i-1}], s^t),\\
 R^t &= \mathrm{ReadPointer}([c_1, c_2, \cdots, c_l], s^t), \\
 W^t &= \mathrm{WritePointer}([c_1, c_2, \cdots, c_l], s^t),
 \label{eq:memory2}
\end{align}
where $\mathcal{M}^t, R^t, W^t$ denote the chosen module, the read pointers and write pointers at time $t$. $\mathrm{ModuleSelect}$, $\mathrm{Read}$, and $\mathrm{Write}$ are pointer functions described in Pointer Networks \cite{Vinyals2015PointerN}. Practically, there are two pairs of read pointers and one pair of write pointers specifying start and end positions of reading and writing. Additionally, Positional Embedding \cite{Vaswani2017AttentionIA} is combined with character embedding to provide the model with relative positional information.

Then the read pointer $R^t$ reads a sub-expression $\hat{e}_{1:p}^t$ from the memory and sends $\hat{e}_{1:p}^t$ to the selected module $\mathcal{M}^t$.
\begin{align}
 \hat{e}_{1:p}^t &= \mathrm{Read}([c_1, c_2, \cdots, c_l], R^t),\\
 \hat{c}_{1:q}^t &= \hat{M}^t(\hat{e}_{1:p}^t),
\label{eq:memory}
\end{align}
where $\hat{c}_{1:q}^t$ is output of module $\mathcal{M}^t$, which is further written into memory.
\begin{align}
 (c_1^{t+1}, c_2^{t+1}, \cdots, c_l^{t+1}) &= \mathrm{Write}(\hat{c}_{1:q}^t, W^t).
\end{align}

\subsection{Optimization}

When the ISM generates the whole actions trajectory $\tau = (\mathcal{M}^1, R^1, W^1,\mathcal{M}^2, R^2, W^2,$ $ \cdots, \mathcal{M}^T, R^T, W^T)$, where $T$ number of the select skill modules, it can output an answer.

Finally, the ISM gets reward 1 when it gives the correct answer. If not, the reward is negative, based on character-level similarity to the solution.

Among reinforcement learning methods, Proximal Policy Optimization (PPO) \cite{schulman2017proximal} is an online policy gradient approach that achieves state-of-the-art on many benchmark tasks. Therefore, we implement PPO to train ISMs. We sample from policies $\pi_{\theta}$ where $\theta$ denotes model parameters. With every state $s$ and sampled action $a=\{\mathcal{M}, R, W\}$, we compute gradients to maximize the following objective function:
\begin{align}
O(\theta) & = O_{PPO}(\theta, \hat{A}) + \alpha * H(\pi_{\theta}(a|s)),
\end{align}
where $\hat{A}$ is the advantage function representing the discounted reward, $H(\cdot)$ is entropy regularizer to discourage premature convergence to suboptimal policies \cite{mnih2016asynchronous}, and $\alpha$ is the coefficient to balance the Exploration-Exploitation, which will be mentioned in CTCS framework (see Section \ref{sec:curri}).

\subsection{Curriculum Teacher and Continual-learning Student (CTCS)}\label{sec:curri}

We propose Curriculum Teacher and Continual-learning Student (CTCS) framework to help the model acquire knowledge efficiently.

\begin{figure}[t]
\centering
\includegraphics[width=0.75\linewidth]{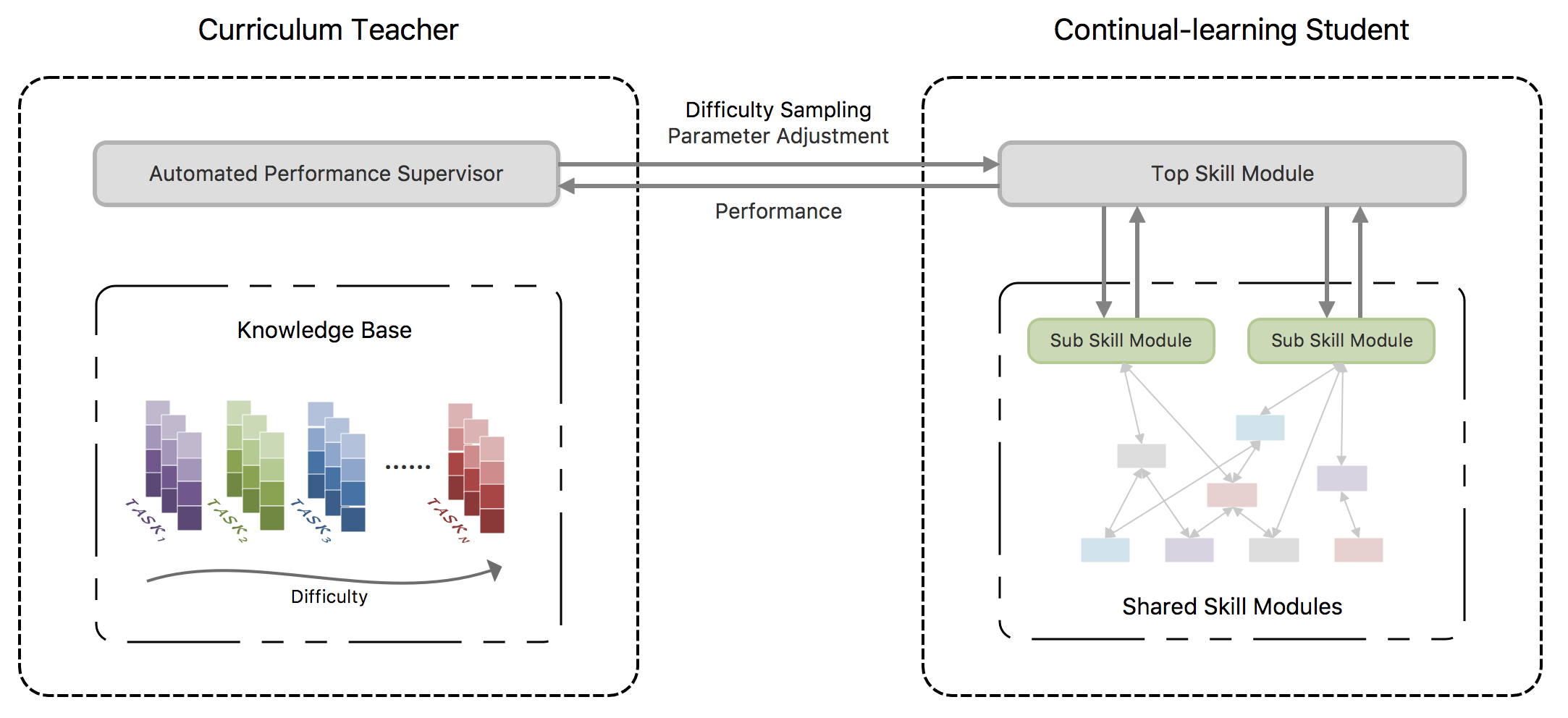}
\caption{Curriculum Teacher and Continual-learning Student (CTCS) framework.}\label{fig:ctcs}
\end{figure}

The CTCS framework is illustrated in Figure \ref{fig:ctcs}. Given a set of $N$ tasks $\{t_1, t_2, \cdots, t_{N}\}$ that are ordered by increasing difficulty. Each task contains $M$ data samples: $t_i = \{x_1^{(i)}, x_2^{(i)}, \cdots, x_{M}^{(i)}\}$). The curriculum teacher gives tasks in the order of $1, 2, \cdots, N$, switching to the next task only when the student performs perfectly in the current task. In learning every task, the curriculum teacher uses difficulty sampling strategy to sample from $M$ data samples.

\textbf{Difficulty Sampling} encourages learning difficult samples. Unlike most problems, arithmetic learning needs precise calculation, which requires complete mastery of training samples. However, the model tends to gain good performance, but not perfect scores. Inspired by Deliberate Practice, a common learning method for human beings, we use difficulty sampling to help the student achieve complete mastery.

To formalize, a difficulty score $d_j^{(i)}$ is the total number of incorrect attempts of sample $j$. Then the probability of each sample $p_{j}^{(i)}$ is determined by a parameterized $Softmax$ function:
\begin{equation}
p_{j}^{(i)} = {\frac {\exp(d_{j}^{(i)}/\tau)}{\sum _{k=1}^{M}\exp(d_{k}^{(i)}/ \tau )}}.
\label{eq:normalize}
\end{equation}

\textbf{Parameter Adjustment} encourages or discourages the exploration of the student. In reinforcement learning, adding entropy controlled by a coefficient $\alpha$ to loss is a commonly used technique \cite{mnih2016asynchronous} to discourage premature convergence to suboptimal policies. However, to what extent should we encourage the student to explore is a long-standing issue of Exploration-Exploitation Dilemma \cite{kaelbling1996reinforcement}. Intuitively, exploration should be encouraged when the student has difficulty doing some samples. Therefore, we employ the teacher to help the student change exploration strategy in keeping with its performance. To be specific, the entropy coefficient $\alpha$ is:
\begin{equation}
\alpha = \min(\beta , \gamma * \max_{j}{d_j^{(i)}})
,\label{eq:entropy}
\end{equation}
where $d_j^{(i)}$ is the difficulty score described in difficulty sampling, $\beta = 0.5$ and $\gamma = 0.01$. As shown in Section \ref{sec:discussion}, difficulty sampling and parameter adjustment methods are critical in achieving the perfect performance.

\section{Experiments}

\paragraph{Arithmetic Expression Calculation}

To train our model to calculate arithmetic expressions with curriculum learning, we define several sub-tasks, from basic tasks like the single-digit addition to compositional tasks like multi-digit division. Then we train our model with tasks in the order of increasing difficulty. The full curriculum list is shown in the appendix. The code is available here\footnote{The code is at (Anonymous).}.

The arithmetic expression data is generated through a random process. An expression of length 10 contains approximately 3 arithmetic operators of $\{+, -, \times, \div, (, )\}$ in average.

We compare our model with two baseline models:

\begin{itemize}
\item Seq2seq LSTM: A sequence to sequence model \cite{sutskever2014sequence} with LSTM \cite{Hochreiter1997LongSM} as encoder and decoder.
\item Neural GPU: An arithmetic algorithm learning model proposed by \citet{kaiser2015neural}. We use their open source implementation posted on Github.
\end{itemize}

\begin{figure}[th]
\centering
\includegraphics[width=0.9\linewidth]{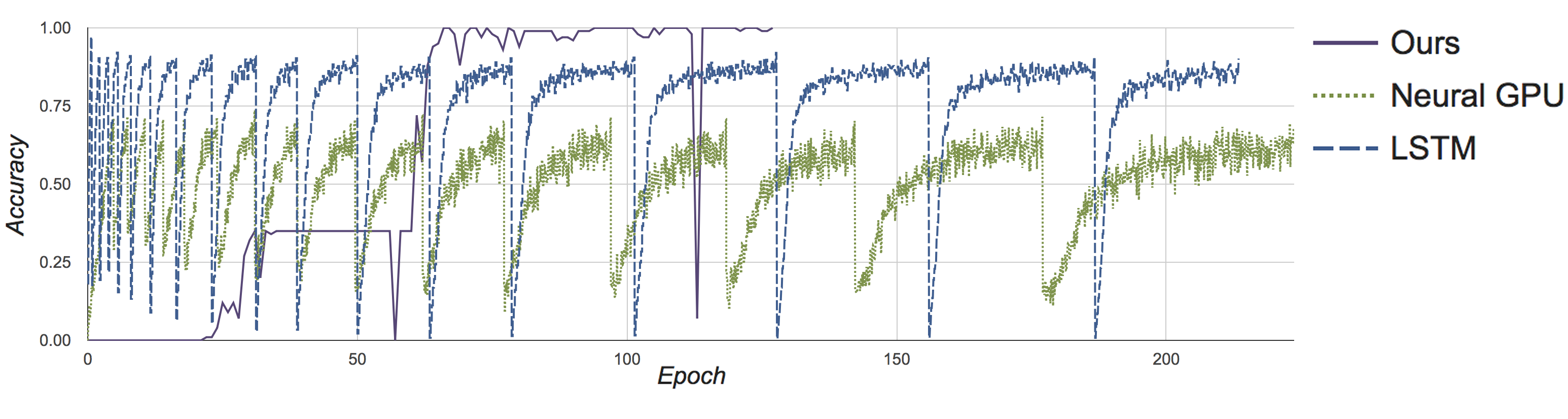}
\caption{Accuracy in training set of size 1000. Training set is changing from easy tasks to difficult tasks as curriculum learning is applied. Every sudden drop indicates a task switching.}\label{fig:train}
\end{figure}

\begin{figure}[th]
\centering
\includegraphics[width=0.9\linewidth]{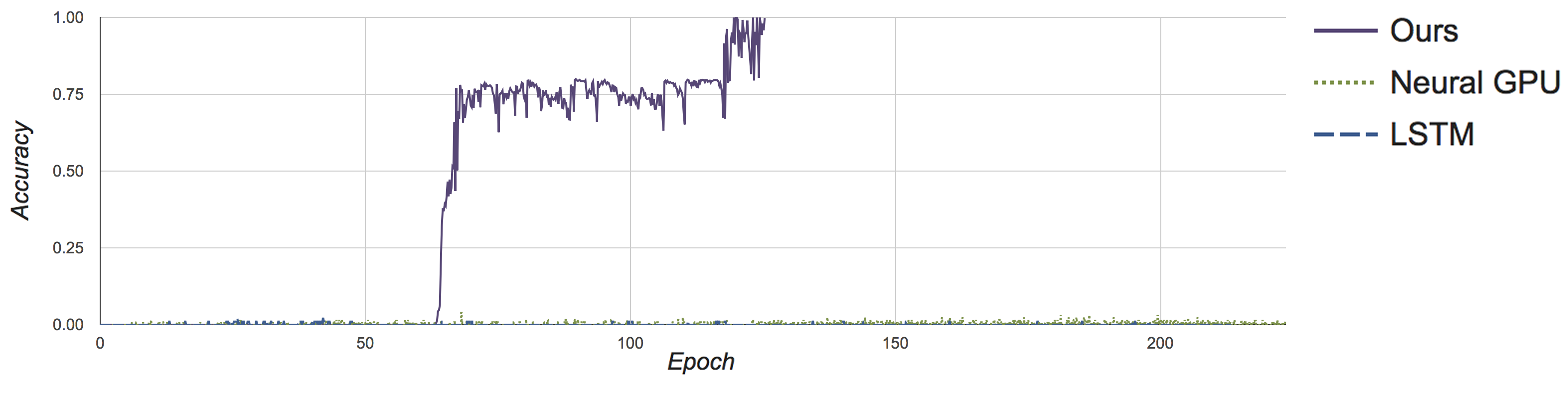}
\caption{Accuracy in test set of size 1000. Test set contains arithmetic expressions of length 10. As the figure shows, the accuracy of both Neural GPU and LSTM is constantly nearly zero.}\label{fig:test}
\end{figure}

To make an objective comparison, we also apply the same curriculum learning method to baseline models. The results are shown in Figure \ref{fig:train} and Figure \ref{fig:test}.

\begin{table}[th]\small
\centering
\subfloat[][Addition.]{
\begin{tabular}{llll}
\toprule
Length  & 5  & 10 & 20 \\
\midrule
Ours       & \textbf{100\%}  & \textbf{100\%}  &  17\% \\
Neural GPU & \textbf{100\%}  & 98\%   & \textbf{84\%} \\
LSTM       & 61\%   & 33\%   & 16\% \\
\bottomrule
\end{tabular}}
\qquad
\subfloat[][Substraction.]{
\begin{tabular}{llll}
\toprule
Length  & 5  & 10 & 20 \\
\midrule
Ours       & \textbf{100\%}  & \textbf{100\%}  &  19\% \\
Neural GPU & \textbf{100\%}  & 72\%   & \textbf{43\%} \\
LSTM       & 95\%   & 48\%   & 20\% \\
\bottomrule
\end{tabular}}
\qquad
\subfloat[][Multiplication.]{
\begin{tabular}{llll}
\toprule
Length  & 5  & 10 & 20 \\
\midrule
Ours       & \textbf{100\%}  & \textbf{100\%}  &  0\% \\
Neural GPU & 30\%   & 3\%    &  0\% \\
LSTM       & 12\%   & 3\%    & 0\% \\
\bottomrule
\end{tabular}}
\qquad
\subfloat[][Division.]{
\begin{tabular}{llll}
\toprule
Length  & 5  & 10 & 20 \\
\midrule
Ours       & \textbf{100\%}  & 27\%  &  15\% \\
Neural GPU & 30\%   & 29\%    &  21\% \\
LSTM       & 28\%   & 23\%    & 19\% \\
\bottomrule
\end{tabular}}
\qquad
\subfloat[][Expression calculation.]{
\begin{tabular}{llll}
\toprule
Length  & 5  & 10 & 20 \\
\midrule
Ours       & \textbf{100\%}  & \textbf{100\%}  &  \textbf{78\%} \\
Neural GPU & 48\%   & 2\%    &  0\% \\
LSTM       & 5\%   & 0\%    & 0\% \\
\bottomrule
\end{tabular}}
\vspace{1em}
\caption{Results of addition, subtraction, multiplication, division, and expression calculation at different length of arithmetic expression in the test set. Task expressions contain one specific arithmetic operator except for the expression calculation task.}
\label{tab:per}
\end{table}

As the result shows, both the baseline models are striving to remember training samples, achieving relatively high accuracy in the training set, but nearly zero accuracy in the test set.
The LSTM model shows a powerful capability of remembering training samples. Every time the task switches, the performance suddenly drops down to zero and then increases to a high level. Although the Neural GPU seems to have better generalization ability, it still performs poorly in the test set.

In contrast, our model achieves almost 100 percent correctness in the experiment, which shows the effectiveness and generalization ability of our model.

\paragraph{Sub-task Performance}

We evaluate our model with different sub-tasks to see the performance of various arithmetic operations. The results are shown in Table \ref{tab:per}. It's noteworthy that the answer of the division is relatively small, so the models can guess the answer, resulting in nearly 20\% correctness in the division. As the result shows, our model achieves 100\% mastery much more than baseline models, especially in expression calculation task.


\paragraph{Hyperparameters}
The gradient-based optimization is performed using the Adam update rule \cite{kingma2014adam}. Every RNN in our model is GRU \cite{chung2014empirical} with hidden size 100. $\tau$ used in Equation \ref{eq:normalize} is 10. The consecutively sample number $N_c$ described in difficulty sampling is 64. In PPO, the reward discount parameter is 0.99, and the clipping parameter $\epsilon$ is 0.2.

\section{Discussion}\label{sec:discussion}

\subsection{Ablation Study}
\paragraph{Curriculum Learning and Continual Learning}

\begin{figure}[h]
\centering
\includegraphics[width=0.78\linewidth]{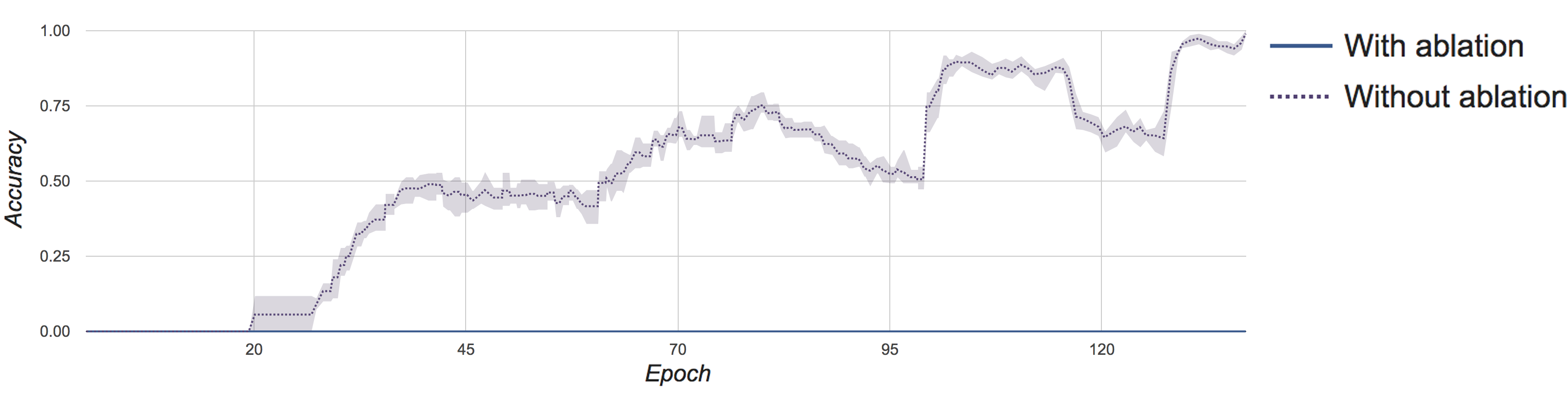}
\caption{Results for ablation of curriculum learning and continual learning. The test set contains arithmetic expressions of length 10 with the modular operator. }\label{fig:curriculum}
\end{figure}

To test if our model can make use of prior knowledge when meeting a new task, we challenge our model with learning a new arithmetic operation: modular. We compare our proposed model with a baseline model that learns from scratch. The results are shown in Figure \ref{fig:curriculum}. Without curriculum learning and continual learning, the model fails to give any correct solutions. It shows the necessity of curriculum learning and continual learning.

\begin{figure}[H]
\centering
\includegraphics[width=0.78\linewidth]{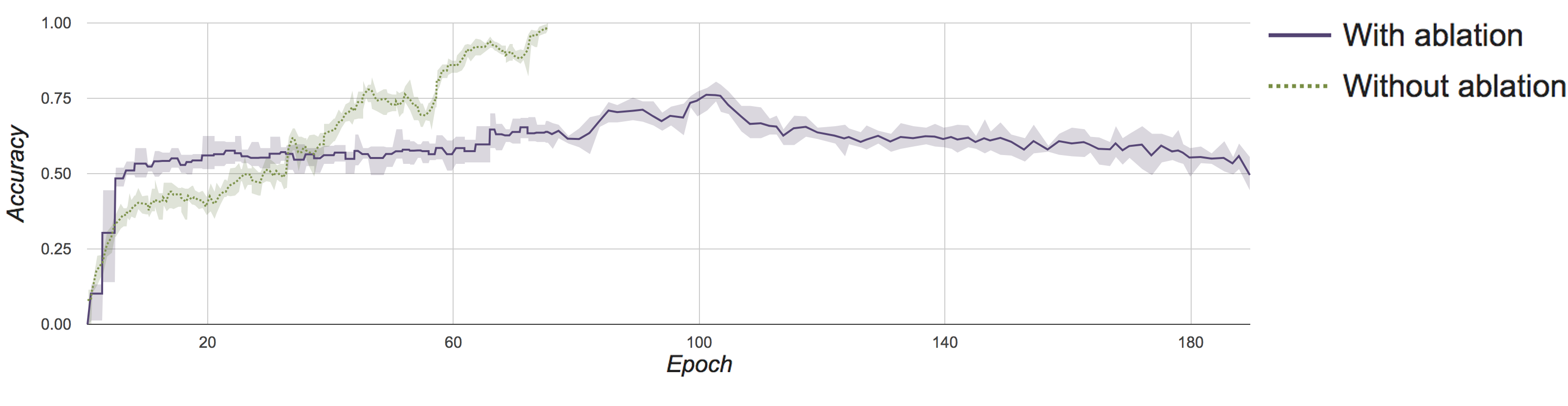}
\caption{Results for ablation of Difficulty Sampling and Parameter Adjustment. The test set contains arithmetic expressions of length 10.}\label{fig:sampling}
\end{figure}

\paragraph{Difficulty Sampling and Parameter Adjustment}
In Curriculum Teacher Continual-learning Student (CTCS) framework, we present difficulty sampling and parameter adjustment to help the model produce the perfect performance. The effectiveness of them is illustrated in Figure \ref{fig:sampling}. Without difficulty sampling and parameter adjustment, the model shows convergence in suboptimal strategy. It shows that difficulty sampling and parameter adjustment are important in helping the model to achieve perfect mastery.

\section{Conclusion}

In this paper, we propose a pure neural model to solve the arithmetic expression calculation problem. Specifically, we use the Multi-level Hierarchical Reinforcement Learning (MHRL) framework to factorize a complex arithmetic operation into several simpler operations. 
We also present Curriculum Teacher Continual-learning Student (CTCS) framework where the teacher adopts difficulty sampling and parameter adjustment strategies to supervise the student. All these above contribute to solving the arithmetic expression calculation problem. Experiments show that our model significantly outperforms previous methods for arithmetic expression calculation.

\bibliographystyle{unsrtnat}
\bibliography{nips_2018}
\end{document}